%% file: root.tex
\documentclass[conference]{IEEEtran}
\IEEEoverridecommandlockouts

\usepackage{cite}
\usepackage{amsmath,amssymb,amsfonts}
\usepackage{graphicx}
\usepackage{textcomp}
\usepackage{xcolor}

\usepackage{algorithm}
\usepackage{algpseudocode}
\usepackage{booktabs}
\usepackage{makecell}
\usepackage{balance}
\usepackage{stfloats}

\usepackage{indentfirst}
\usepackage{multirow}
\usepackage{rotating}
\usepackage{array}
\usepackage{tabularx}
\newcolumntype{C}[1]{>{\centering\arraybackslash}m{#1}}
\newcolumntype{Y}{>{\centering\arraybackslash}X}

\usepackage{bm}
\usepackage{pifont}

\usepackage{xcolor}
\usepackage{soul}
\usepackage{float}
\usepackage{url}
\usepackage{graphicx}
\usepackage{subcaption}
\usepackage[caption=false,font=footnotesize]{subfig}

\def\BibTeX{{\rm B\kern-.05em{\sc i\kern-.025em b}\kern-.08em
    T\kern-.1667em\lower.7ex\hbox{E}\kern-.125emX}}

\begin{document}

\setlength{\textfloatsep}{4pt}
\setlength{\intextsep}{4pt}
\setlength{\floatsep}{4pt}

\title{INSIGHT: Enhancing Autonomous Driving Safety through Vision-Language Models on Context-Aware Hazard Detection and Reasoning\\
\thanks{This work is supported  NDSU VPR Office project, Accelerating the Deployment of Autonomous Vehicles in Rural Areas, and National Science Foundation under Award SaTC--2350075.}
\thanks{ $\IEEEauthorrefmark{1}$ D. Chen, L. Cheng, and XT. Yang are with the College of Engineering, University of Maryland, College Park, MD 20742, USA. (Email: \{dwchen98, leicheng, xtyang\}@umd.edu).
\textit{(Corresponding author: Xianfeng Terry Yang.)}}
\thanks{
 $\IEEEauthorrefmark{2}$ Z. Zhang and Y. Liu are with the Department of Computer Science, North Carolina State University, Raleigh, NC, 27695, USA (Email: \{zzhang66, yuchen.liu\}@ncsu.edu).
}
}

\author{
		\IEEEauthorblockN{Dianwei Chen $\IEEEauthorrefmark{1}$,
		Zifan Zhang $\IEEEauthorrefmark{2}$,
        Lei Cheng $\IEEEauthorrefmark{1}$
        Yuchen Liu $\IEEEauthorrefmark{2}$,
        Xianfeng Terry Yang $\IEEEauthorrefmark{1}$
 }
	}
\maketitle
\maketitle
\begin{abstract}

Ensuring safety in autonomous driving requires not only accurate perception of surrounding agents and infrastructure but also a human-like understanding of rare, safety-critical edge cases. This paper presents INSIGHT, a hierarchical vision–language model (VLM) framework for context-aware hazard detection and reasoning in autonomous driving scenes. Built on Qwen2-VL, INSIGHT introduces a commonsense-constrained supervised fine-tuning strategy that fuses human risk priors with multimodal inputs to guide the model’s attention toward potentially hazardous regions. A lightweight 2D heatmap head with a differentiable soft-argmax is attached to the backbone, enabling joint optimization of narrative sufficiency and spatially grounded coordinate regression through a multi-task loss. To support training and evaluation, we curate a 1,000-sample hazard-awareness dataset from BDD100K, where each image is labeled with a single human-annotated hazard point and a corresponding natural-language rationale. Quantitative results on this benchmark demonstrate that INSIGHT substantially improves BLEU, ROUGE, and pixel-level localization error compared with Qwen2-VL baselines, while qualitative attention visualizations show sharper, risk-aware focus on critical objects and regions in both urban and highway scenarios. Experimental results confirm significant gains over existing VLM baselines in both area prediction accuracy and semantic relevance, bringing autonomous vehicles one step closer to human-level situational awareness.

\end{abstract}

\vspace{+0.1cm}
\begin{IEEEkeywords}
Vision-language model, autonomous driving, safety-critical scenario, hazard detection
\end{IEEEkeywords}

\input{sec/intro}

\input{sec/related}
\input{sec/method}
\input{sec/exp}

\input{sec/conclusion}



\bibliographystyle{IEEEtran}
\bibliography{root}

\end{document}

%% file: sec/intro.tex
\section{Introduction}




Ensuring safety in autonomous driving hinges on the vehicle’s ability to perceive, understand, and react to complex real‑world environments in real time, with perception serving as the foundational prerequisite~\cite{teng2025improving}. Current perception stacks are typically built with separate modules for sensor fusion, detection, and prediction. They perform well in common scenarios but often fail to reason about rare, fast-evolving, or even unhappened but potential safety-critical edge cases that ultimately cause most safety-critical incidents~\cite{tian2024drivevlm}.
Additionally, existing stacks lack the contextual understanding and semantic reasoning needed to recognize subtle cues or anticipate atypical situations before they escalate into hazards. To address this gap, we explore the use of recent vision–language models (VLMs), which offer a powerful capability to jointly process visual inputs and linguistic prompts. By aligning pixels with words, VLMs provide a unified semantic representation of the driving scene that is both machine-operable and human-interpretable~\cite{ng2025vision}. This modality-bridging capability enables context-aware hazard detection and proactive edge case reasoning, even in zero-shot or low-shot scenarios, where traditional models fail to generalize~\cite{yang2024hard}. Compared to conventional computer vision and rule-based reasoning systems, VLMs offer greater flexibility, richer interpretability, and seamless multi-task integration, making them a promising foundation for enhancing autonomous driving safety.

More specifically, we focus on a critical limitation of current detection and prediction models in autonomous driving: limited alignment with human risk perception. While modern perception systems achieve strong performance on common scenarios, they often struggle in subtle yet safety-critical situations that are intuitively obvious to human drivers. This gap hinders the system’s ability to anticipate potential hazards before they fully manifest.

To overcome this challenge, we propose \textbf{INSIGHT}, a hierarchical VLM framework designed for context-aware hazard detection and reasoning. Unlike conventional hazard detection models and existing VLM pipelines, our method introduces three key innovations:

\begin{itemize}
    \item Introduces a commonsense-constrained supervised fine-tuned (SFT) method that fuses human sense knowledge with multimodal data to guide VLM attention toward potential and actionable hazards;  
    \item Proposes a unified textual–visual framework that combines LLMs' prior knowledge with human interpretability for scenario risk area prediction and reorganization;  
    \item Formulates a dual‑task loss that jointly optimizes coordinate regression and narrative sufficiency, delivering state‑of‑the‑art hazard localization accuracy on BDD100K driving scenario dataset safety-critical scenario splits;  
\end{itemize}

By bridging high‑level semantic reasoning with low‑level spatial precision, INSIGHT pushes VLM‑based perception beyond passive semantic generation toward an active safety subsystem that anticipates, explains, and mitigates on‑road hazards. 


%% file: sec/related.tex
\section{Related Works}

\begin{figure*}[htbp]
    \centering
    \includegraphics[width=1\linewidth]{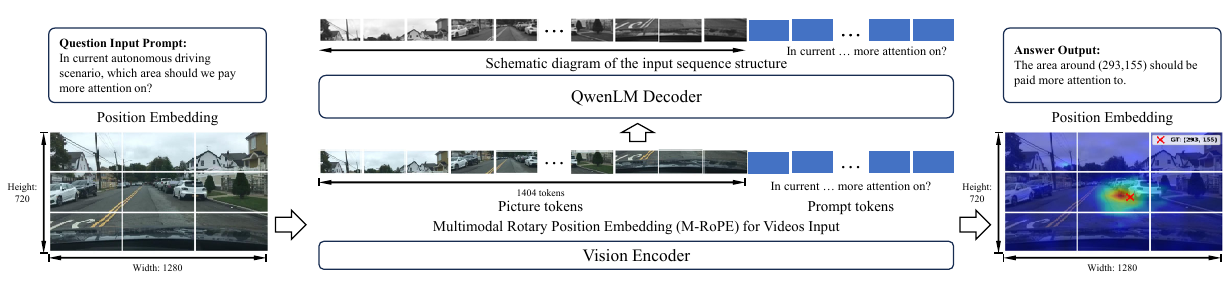}
    \caption{Inference framework of INSIGHT SFT-finetuned Qwen-VL model}
    \label{fig:framework}
\end{figure*}

Recently, the development of autonomous driving algorithms has significantly improved with various approaches addressing scene understanding, prediction, and control~\cite{min2024driveworld,zhao2024enhanced,huang2020multi}. This section reviews relevant work in safety-critical scenario exploration, end-to-end autonomous driving models, and VLMs in autonomous vehicles.
\subsection{Safety-Critical Scenario Exploration}
Safety-critical scenarios in autonomous driving refer to situations that pose elevated risks to traffic participants and require timely and accurate perception, prediction, and decision-making to prevent potential accidents~\cite{Cai2024A, Karunakaran2023Generating}. Verifying both the vehicle and its algorithms in these scenarios is critical to ensuring safety and reliability~\cite{feng2023dense}. 
Testing in such scenarios involves subjecting the vehicle and its systems to high-risk or failure-prone conditions, allowing for a comprehensive evaluation of their performance under adverse circumstances~\cite{goss2021generation, Zhang2023Finding}.
These simulations play a vital role in rigorously testing and refining algorithms and vehicle systems, enabling the identification and resolution of potential weaknesses~\cite{Saffary2024Developing}. After verifying and even training in such scenarios, these vehicles are able to not only enhance the safety assurance and reliability of autonomous systems but also accelerate their deployment in real-world settings by ensuring they meet safety and performance standards under a wide range of conditions 
%
Extensive research has been conducted to model and simulate safety-critical scenarios, focusing on conditions such as extreme weather, erratic pedestrian behavior, and unconventional vehicle movements~\cite{chen2023using, Luo2021Interactive, chen2024deep, chen2025advanced}. By addressing these issues in simulated environments, developers can ensure a higher level of safety and performance before deploying the systems in real-world applications~\cite{Ghodsi2021Generating}. This proactive approach is essential for building trust in autonomous driving technologies and mitigating risks associated with safety-critical scenarios.

\subsection{End-to-End Autonomous Driving Model}
End-to-end models for autonomous driving utilize deep learning architectures to map raw sensor inputs, such as images and LiDAR data, directly to control outputs like steering, acceleration, and braking. Convolutional neural network (CNN) is a common model employed to process visual data by extracting spatial features from camera inputs, enabling the recognition of lanes, vehicles, and other road elements~\cite{sharma2023cnn, alsanwy2023cnn}. Recurrent neural network (RNN) and temporal convolutional network (TCN) handle data sequences effectively, capturing dynamic changes in the environment over time~\cite{Du2021Imitation}. Moreover, transformer utilizes self-attention mechanisms to effectively model long-range dependencies and complex interactions for both visual and sequential data processing~\cite{rayakota2024hybridte, chen2023detrive}, offering improved performance in tasks such as object detection, trajectory prediction, and scene understanding~\cite{Li2023Lane}. While effective in many scenarios, these models often face challenges in rare or hard-to-predict safety-critical scenarios due to their reliance on the data used during training, which is collected from daily common driving scenarios or simple AV simulator scenarios~\cite{10605806}, which may not fully represent real-world variability.

\subsection{Vision-Language Model in Autonomous Vehicle}
In recent years, VLM has made breakthroughs in natural language processing and multimodal tasks~\cite{cui2023drivellm}. Applying it to autonomous driving can effectively improve scene understanding and decision-making capabilities~\cite{zhou2024vision, Pan2024VLP}.
VLMs enhance the system’s comprehensive perception capabilities by combining visual data, such as cameras and LiDAR, and textual data, such as traffic signs and navigation instructions~\cite{zheng2024simplellm4ad,zhang2022learning}. The effective integration of visual features and language representation through visual language adapters can improve the ability to understand complex driving scenarios~\cite{xu2024drivegpt4}.
With large-scale pre-trained VLMs, the model can learn common representations and knowledge from massive multi-modal data, assisted with zero-shot learning~\cite{wang2024solving}, thus having strong generalization capabilities~\cite{ Wei2022Semantic}. The model can infer semantic information, such as the meaning of the potential movement of a pulled-over vehicle with law enforcement on the side, from visual features in an image~\cite{Su2024To}. This ability allows the model to use pre-trained knowledge to make inferences about unknown data when encountering new scenarios, without relying on task-specific annotated data~\cite{gouidis2024fusing}.
In our work, we use VLM to unify visual and language understanding in autonomous driving, enabling real-time decisions, better scene comprehension, and improved edge case handling for safer and more efficient driving automation.

%% file: sec/method.tex
\section{INSIGHT}
\label{sec:insight}
\subsection{INSIGHT Framwork Overview }
As shown in Fig.~\ref{fig:framework}, the INSIGHT inference framework is built on a Qwen-VL model finely tuned by SFT. First, on the visual side, video frames from the autonomous driving scene are extracted into a series of picture tokens by the Vision Encoder, and spatial and temporal location information is incorporated through Multimodal Rotary Position Embedding (M-RoPE) to form a multimodal representation suitable for video input. On the text side, natural language questions such as "Which areas should be focused on in the current autonomous driving scene?" are encoded into prompt tokens, and location encoding is also added. Subsequently, these two sequences are concatenated in a unified input structure and fed into the QwenLM Decoder to achieve joint modeling of scene semantics and spatiotemporal cues. The decoder output not only provides the coordinates of the areas requiring focus (e.g., (293, 155)), but can also be further mapped into a heatmap visualization, thus intuitively demonstrating the model's ability to identify potential risk areas in complex road environments.

\begin{figure}
    \centering
    \includegraphics[width=1\linewidth]{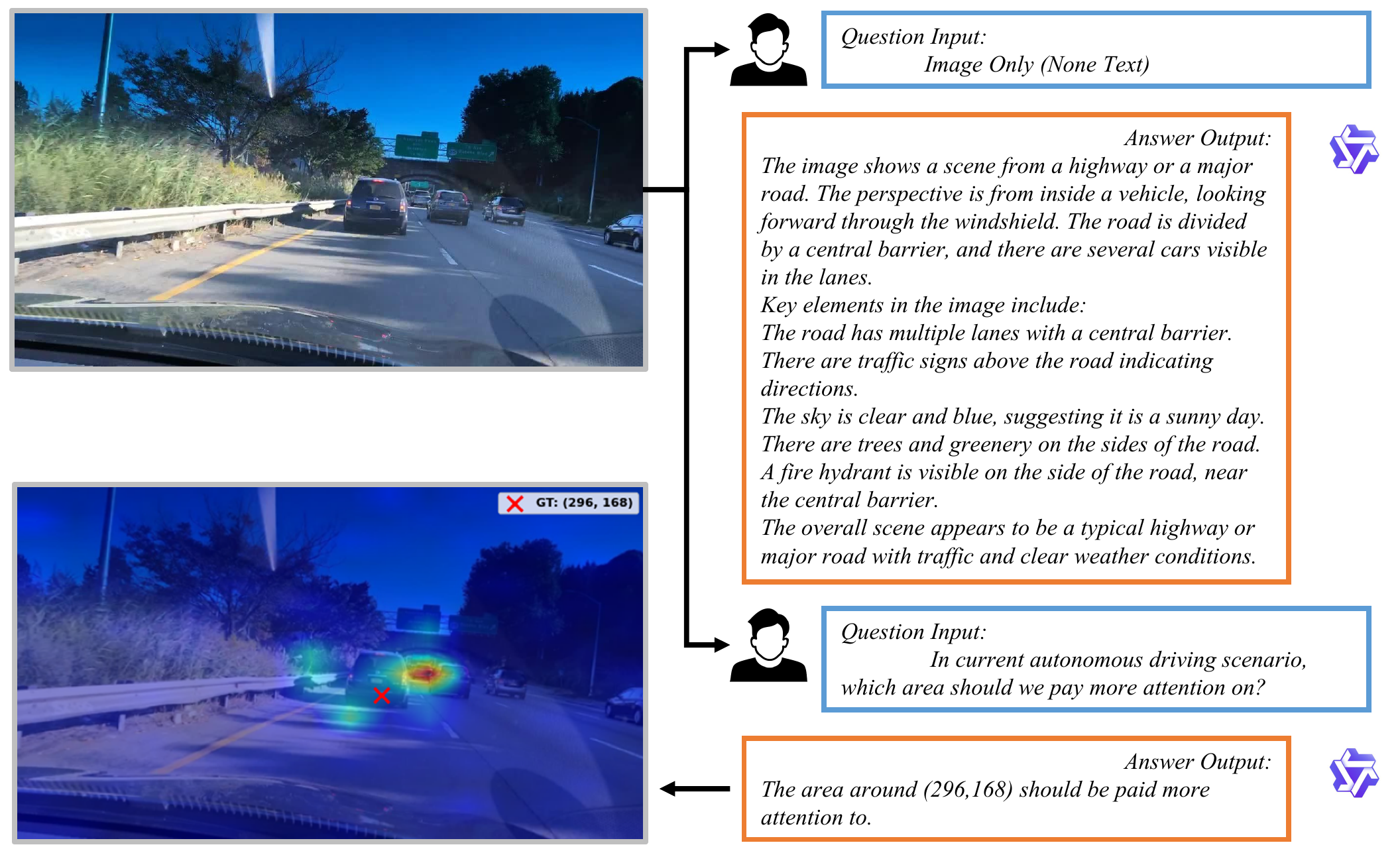}
    \caption{Generalization ability demonstration}
    \label{fig:placeholder}
\end{figure}

The baseline VLM (e.g., Qwen2-VL) setup follows the standard recipe: it optimizes a single language–modeling loss $\mathcal{L}_{\text{text}}$ on image–conditioned text without any explicit grounding. 
\textbf{Our proposed method} converts this into a grounding–aware multi–task objective by attaching a lightweight \emph{learned 2-D heatmap head} (detailed below) and jointly optimizing
\begin{equation}
\mathcal{L}_{\text{total}} \;=\; \lambda_{\text{text}}\,\mathcal{L}_{\text{text}} \;+\; \lambda_{\text{coord}}\,\mathcal{L}_{\text{coord}},
\label{eq:multi_task_loss}
\end{equation}
where ground–truth points $(x,y)$ are extracted from the annotation text and normalized to $[0,1]^2$ using the original image width and height. 
The heatmap head maps pooled vision–language features to an $H{\times}W$ spatial probability map, and predicted coordinates are obtained via a differentiable \emph{soft-argmax}. 
For efficiency, we fine-tune the base VLM in 4-bit quantized form and apply LoRA to the attention projection modules (q/k/v/o), together with gradient checkpointing. 
We use text cross-entropy and coordinate MSE as training losses; grounding metrics such as PTC/Hit@r are left to future work.

Differences from the baseline VLM(e.g. Qwen2-VL): 

\textit{1) Objective} is to jointly optimize text generation and coordinate regression via the multi-task loss in Eq.~\eqref{eq:multi_task_loss}; 

\textit{2) Supervision} uses single-click point annotations extracted from text and normalized to $[0,1]^2$ to provide lightweight yet informative grounding signals;

\textit{3) Architecture} employs a compact 2-D heatmap prediction head with soft-argmax to output continuous grounded coordinates;

\textit{4) Efficiency} is achieved by applying LoRA to (q/k/v/o) projections along with 4-bit quantization and gradient checkpointing for memory-efficient training.



\subsection{Hallucination Reduction}
Standard VLM fine-tuning assumes answers are implicitly grounded. We make grounding explicit by requiring the model not only to \emph{describe} but also to \emph{point}. The added spatial objective encourages the backbone to focus on causally relevant regions and reduces ``look-and-hallucinate'' failures in traffic scenes, especially for rare or safety-critical scenarios where sparse supervision is most valuable.

Hence, we train the model by minimizing the multi-task loss $\mathcal{L}_{\text{total}}$ in Eq.~\eqref{eq:multi_task_loss} with a fixed $\lambda_{\text{coord}}$ per run. Training inputs are image-conditioned prompts from the dataset (the user message already contains the $<image>$ token); target coordinates are parsed from the assistant message, then normalized by the original image width/height. We train on a subset for efficient iteration and report (i) text loss and (ii) coordinate MSE on validation data.

\subsection{Learned Heatmaps for Spatial Localization}
We employ a transformer-based VLM to produce multimodal embeddings. For spatial grounding, instead of taking an argmax over encoder attentions, we attach a lightweight MLP head that maps a mean-pooled joint vision–language representation to an $H{\times}W$ heatmap; we then apply softmax and obtain coordinates via a differentiable soft-argmax.

\subsubsection{Heatmap and Soft-Argmax Coordinates}
Let $z\in\mathbb{R}^{d}$ be a pooled backbone feature (mean-pooled over tokens). A two-layer MLP produces logits over an $H{\times}W$ grid:
\begin{equation}
\begin{aligned}
\ell &= W_2\,\sigma(W_1 z)\in\mathbb{R}^{HW},\\
A   &= \mathrm{softmax}(\ell)\ \ \text{reshaped to } \mathbb{R}^{H\times W}.
\end{aligned}
\end{equation}

Define normalized grid coordinates $x\in\{0,\tfrac{1}{W-1},\dots,1\}$ and $y\in\{0,\tfrac{1}{H-1},\dots,1\}$. The predicted point is the expectation under $A$:
\begin{equation}
\hat{x} \;=\; \sum_{u,v} x_v\, A_{u,v},\qquad
\hat{y} \;=\; \sum_{u,v} y_u\, A_{u,v},
\end{equation}
yielding continuous, differentiable coordinates $(\hat{x},\hat{y})\in[0,1]^2$.

\subsubsection{Integration with Text Generation}
The language modeling head is trained in parallel on the image-conditioned prompt tokens. The spatial head shares the backbone with the language pathway; thus, improvements in localization can regularize textual predictions and vice versa.

\subsection{Loss Function}
Expanding the terms in Eq.~\eqref{eq:multi_task_loss}, we define
\begin{equation}
\mathcal{L}_{\mathrm{coord}}
= \frac{1}{N}\sum_{i=1}^N
\bigl\|(\hat{x}_i,\hat{y}_i)-(x_i,y_i)\bigr\|_2^2,
\end{equation}
\begin{equation}
\mathcal{L}_{\mathrm{text}} = -\frac{1}{N}\sum_{i=1}^N\sum_{t=1}^T s_{i,t}\log \hat{s}_{i,t},
\end{equation}
where $(x_i,y_i)$ are ground-truth points normalized by the original image size and $(\hat{x}_i,\hat{y}_i)$ come from the soft-argmax over $A_i$.

We use AdamW with a cosine learning-rate schedule and standard HuggingFace Trainer defaults. Low-Rank Adaptation (LoRA) adapters are applied to the attention projections (q/k/v/o). In our sweeps, $\lambda_{\mathrm{coord}}$ is held constant \emph{per run} (no within-run scheduling).
\subsection{Convergence Sketch and Stability Analysis}
We briefly sketch how the multi-task objective in Eq.~\eqref{eq:multi_task_loss} fits into standard AdamW analyses. Let
\begin{equation}
\mathcal{L}_{\mathrm{total}}(\theta)
= \lambda_{\mathrm{coord}}\,\mathcal{L}_{\mathrm{coord}}(\theta)
+ \lambda_{\mathrm{text}}\,\mathcal{L}_{\mathrm{text}}(\theta),
\end{equation}
with $\lambda_{\mathrm{coord}},\lambda_{\mathrm{text}}\ge 0$, and assume:
\subsubsection{Lower-boundedness:} $\mathcal{L}_{\mathrm{total}}(\theta)\ge 0$ for all $\theta$.
\subsubsection{$L$-smoothness:} Each component $\mathcal{L}_i(\theta)$ is Lipschitz-smooth; hence $\mathcal{L}_{\mathrm{total}}$ is Lipschitz-smooth as a nonnegative linear combination of smooth functions. We denote its smoothness constant by $L$.
\subsubsection{Bounded gradients:} There exists $G>0$ such that $\|\nabla \mathcal{L}_{\mathrm{total}}(\theta)\|\le G$ for all $\theta$.

AdamW maintains per-parameter first and second moments:
\begin{equation}
m_{t+1} = \beta_1 m_t + (1-\beta_1) g_t,\qquad
v_{t+1} = \beta_2 v_t + (1-\beta_2) g_t^2,
\end{equation}
with bias corrections $\hat m_{t+1} = m_{t+1}/(1-\beta_1^{t+1})$, $\hat v_{t+1} = v_{t+1}/(1-\beta_2^{t+1})$, where $g_t=\nabla \mathcal{L}_{\mathrm{total}}(\theta_t)$. The update with decoupled weight decay $\lambda_{\mathrm{wd}}$ is
\begin{equation}
\theta_{t+1}
= \theta_t
- \eta_t\, \frac{\hat m_{t+1}}{\sqrt{\hat v_{t+1}} + \epsilon}
- \eta_t\,\lambda_{\mathrm{wd}}\,\theta_t.
\end{equation}
Under standard decaying stepsizes $\sum_{t\ge 0}\eta_t=\infty$, $\sum_{t\ge 0}\eta_t^2<\infty$ and appropriate choices of $(\beta_1,\beta_2)$, existing analyses for Adam-type methods imply a \emph{sublinear} stationarity rate
\begin{equation}
\min_{0\le t<T}\mathbb{E}\bigl\|\nabla \mathcal{L}_{\mathrm{total}}(\theta_t)\bigr\|^2
\le \frac{C}{\sqrt{T}},
\end{equation}
for a constant $C$ depending on $L$, $G$, $\beta_1$, $\beta_2$, and initialization. We do not claim a new convergence theorem here; rather, we note that our multi-task objective satisfies the standard assumptions used in AdamW analyses.

\paragraph{Why the objective is well-behaved.}
The soft-argmax mapping from logits to $(\hat x,\hat y)$ is smooth, so $\mathcal{L}_{\mathrm{coord}}$ inherits smoothness from the backbone (under standard smooth regression losses); the overall multi-task objective remains smooth as a weighted sum. LoRA reduces the number of trainable directions, which often improves empirical stability. In practice we observed stable training under cosine LR schedules and fixed $\lambda_{\mathrm{coord}}$ per run.

\paragraph{Limitations.}
The above is an asymptotic, nonconvex stationarity guarantee under idealized assumptions; rates can degrade with poorly tuned $(\beta_1,\beta_2)$, large $\epsilon$, or extreme class imbalance. A thorough study of $\lambda_{\mathrm{coord}}$ schedules and grounding-specific generalization is left for future work.

\newcommand{\vthumb}[1]{%
  \raisebox{-0.5\height}{\includegraphics[width=\linewidth,height=2.2cm,keepaspectratio]{#1}}%
}

\begin{table*}[ht!]
\centering
\caption{BDD100K Sub-dataset format preview with thumbnails}
\label{tab:hf_view_like_thumb}
\renewcommand{\arraystretch}{1.15}
\begin{tabularx}{0.98\linewidth}{
  >{\centering\arraybackslash}m{0.16\linewidth}
  >{\raggedright\arraybackslash}m{0.12\linewidth}
  X
}
\toprule
\makecell{\textbf{Thumbnails}\\ \textit{(Preview)}} &
\makecell{\textbf{Images}\\ \textit{(Width $\times$ Height)}} &
\makecell{\textbf{Messages}\\ \textit{(List $\cdot$ Lengths)}} \\
\midrule
\vthumb{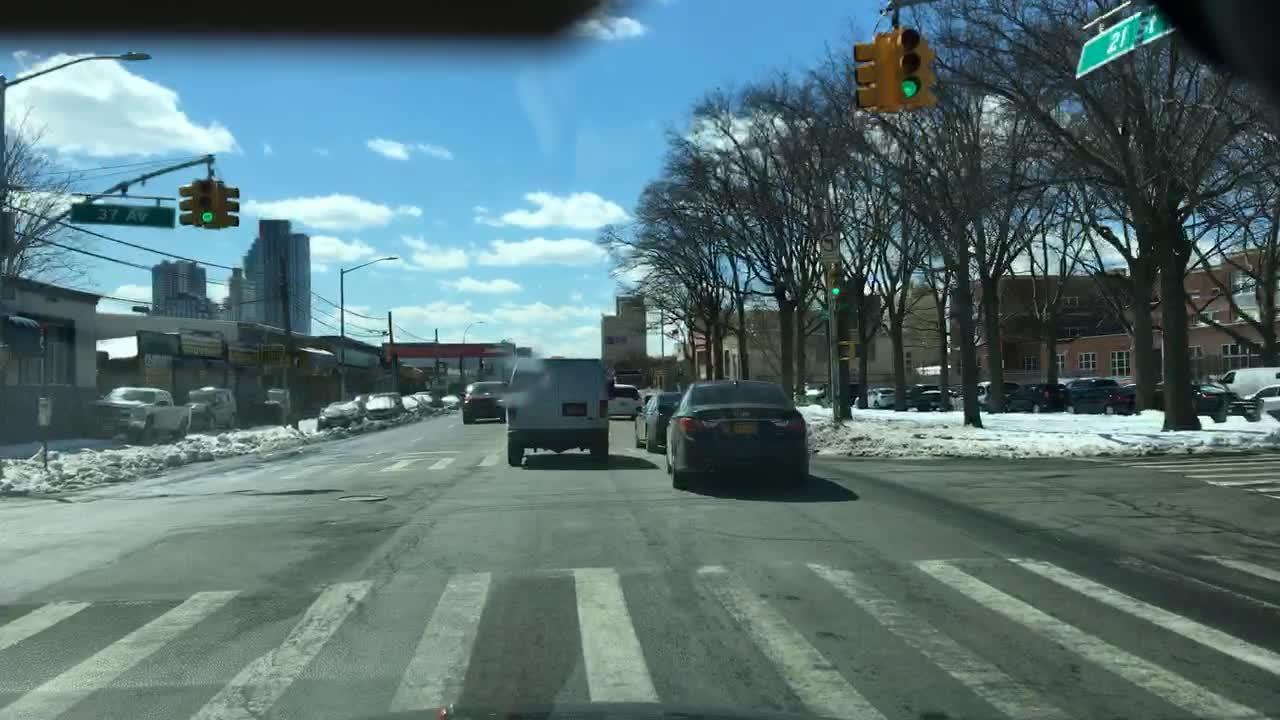} &
\textit{1280 $\times$ 720} \hfill \texttt{RGB} &
\textit{2} \hfill
\texttt{[ \{ "role":"user", "content":"<image> Which area should we pay more attention to in the current autonomous driving scenario?" \}, \{ "role":"assistant", "content":"The area around (544,459) should be paid more attention to." \} ]} \\
\midrule
\vthumb{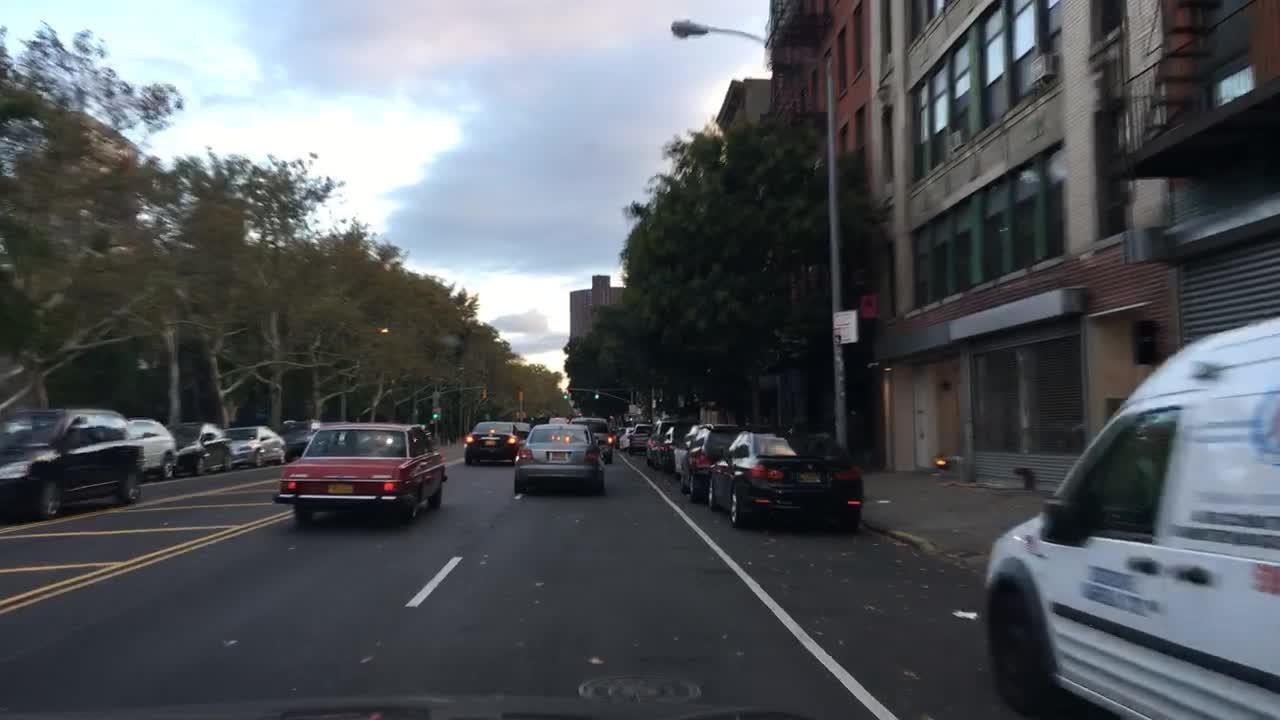} &
\textit{1280 $\times$ 720} \hfill \texttt{RGB} &
\textit{2} \hfill
\texttt{[ \{ "role":"user", "content":"<image> Which area should we pay more attention to in the current autonomous driving scenario?" \}, \{ "role":"assistant", "content":"The area around (376,497) should be paid more attention to." \} ]} \\
\midrule
\multicolumn{3}{c}{\textit{\dots\ 996 rows omitted \dots\ }} \\
\midrule
\vthumb{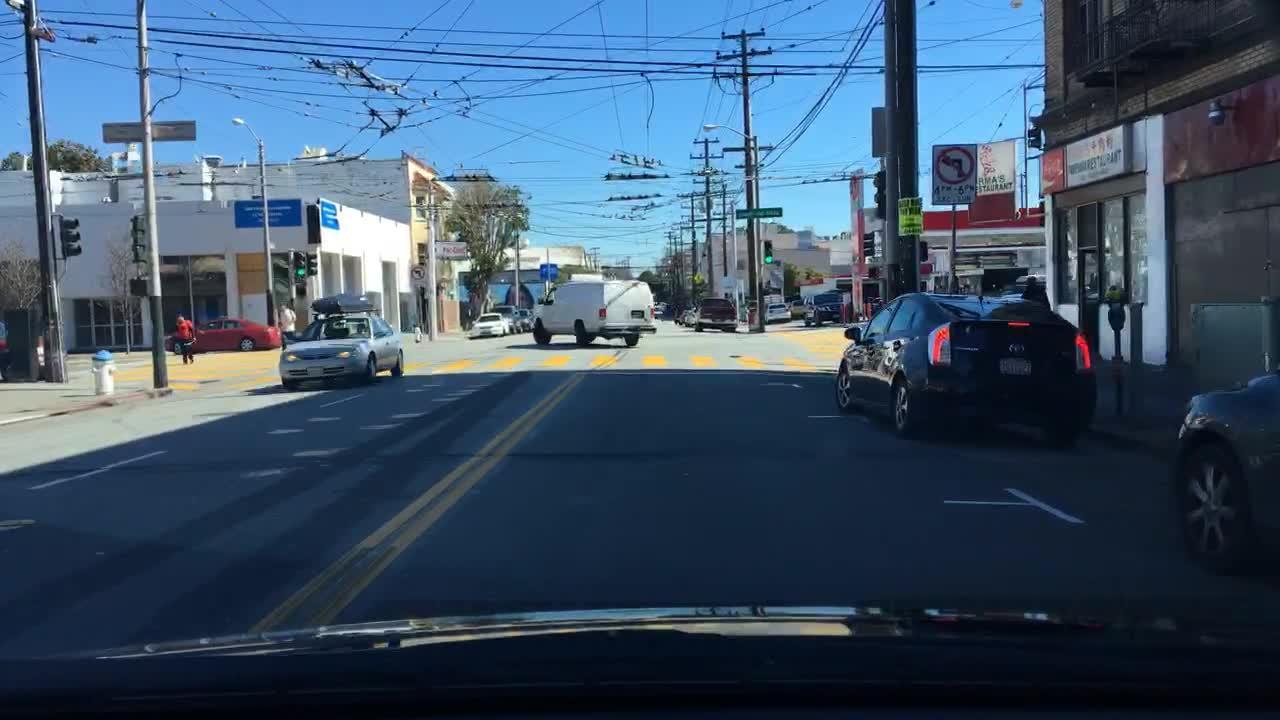} &
\textit{1280 $\times$ 720} \hfill \texttt{RGB} &
\textit{2} \hfill
\texttt{[ \{ "role":"user", "content":"<image> Which area should we pay more attention to in the current autonomous driving scenario?" \}, \{ "role":"assistant", "content":"The area around (866,371) should be paid more attention to." \} ]} \\
\midrule
\vthumb{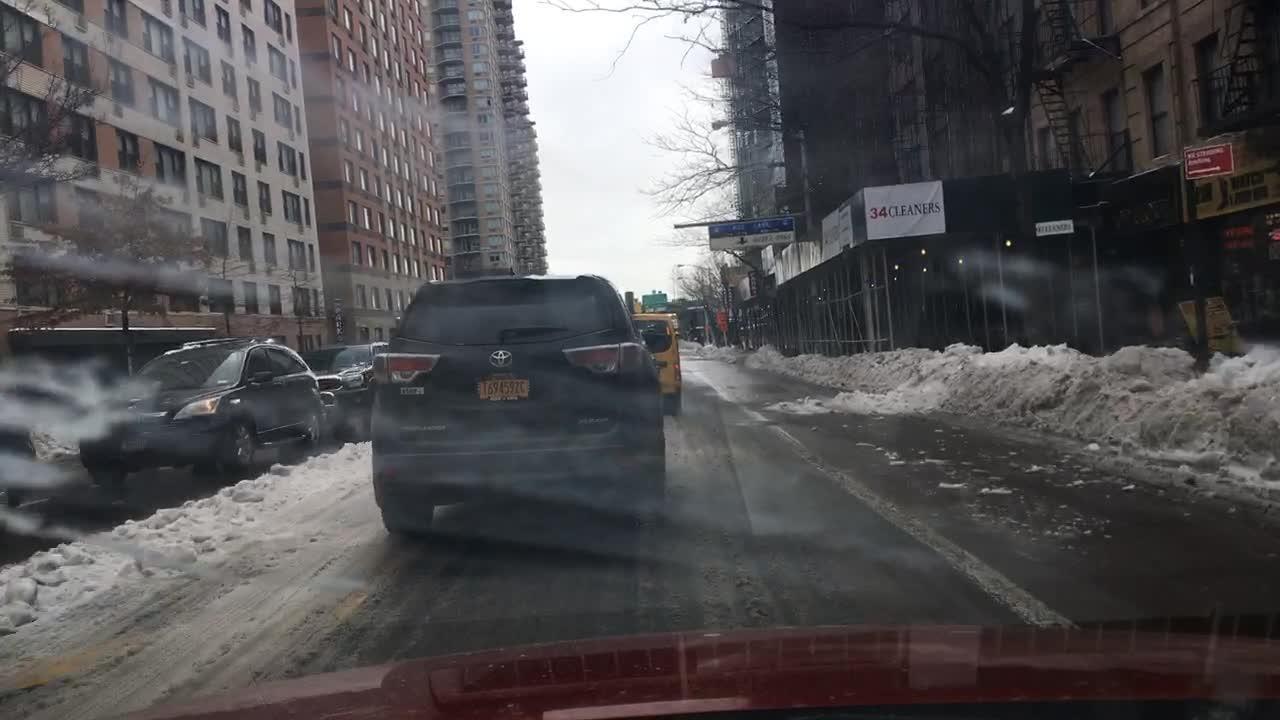} &
\textit{1280 $\times$ 720} \hfill \texttt{RGB} &
\textit{2} \hfill
\texttt{[ \{ "role":"user", "content":"<image> Which area should we pay more attention to in the current autonomous driving scenario?" \}, \{ "role":"assistant", "content":"The area around (497,453) should be paid more attention to." \} ]} \\
\bottomrule
\end{tabularx}
\end{table*}




%% file: sec/exp.tex
\section{Experiments}

\subsection{Dataset Preprocessing}
The dataset used in this study is a curated subset of the BDD100K dataset, a large-scale and diverse driving video corpus containing 100{,}000 videos with rich annotations such as bounding boxes, lane markings, drivable areas, and object tracking, widely used in autonomous driving and computer vision research. The selected images span urban, suburban, and highway scenes under diverse lighting and weather conditions.

\begin{figure}[ht]
    \centering
    \includegraphics[width=0.472\textwidth]{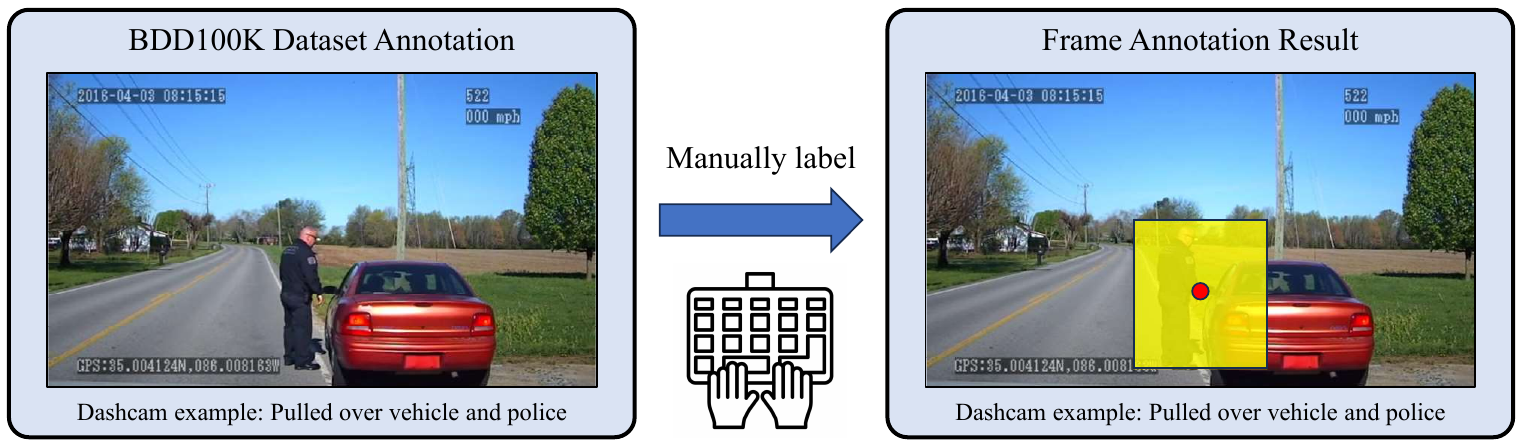}
    \caption{Manual annotation method used for dataset preprocessing.}
    \label{fig:manual_annotation}
\end{figure}

On top of this subset, we construct a custom hazard-awareness dataset by manually annotating each image with potential hazard points \((x, y)\) based on human judgment and driving experience (Fig.~\ref{fig:manual_annotation}).

\subsubsection{Annotation Method}
Manual annotation is employed to label potential hazard areas within each image. A hazard area is defined based on the annotator's driving experience and judgment, covering regions where pedestrians, vehicles, or other obstacles may pose risks to safe driving. In general, these potential hazard areas can be divided into two categories: predictable surrounding behavior (e.g., a vehicle in an adjacent lane attempting to merge) and unpredictable surrounding behavior (e.g., a pedestrian suddenly stepping onto the road).

To balance annotation cost and coverage of key scenarios, we select a subset of 1{,}000 images. During annotation, each frame is displayed for at most 5 seconds, within which the subject identifies the region with the highest probability of becoming hazardous. A bounding box is drawn around this region, and the center of the box is recorded as the hazard point \((x, y)\).

For consistency, a single primary annotator labels all 1{,}000 images, and each image contains exactly one hazard area. To assess the reliability of these annotations, a subset of 100 images is independently reviewed by two additional annotators; disagreements are discussed and resolved, yielding a consensus label set for the entire dataset.

\begin{figure*}[t]
    \centering
    \includegraphics[width=0.95\linewidth]{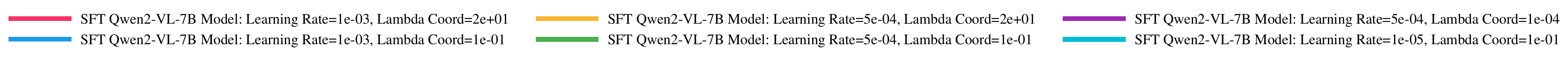}\\[0.5em]

    \begin{minipage}{0.312\linewidth}
        \centering
        \includegraphics[width=\linewidth]{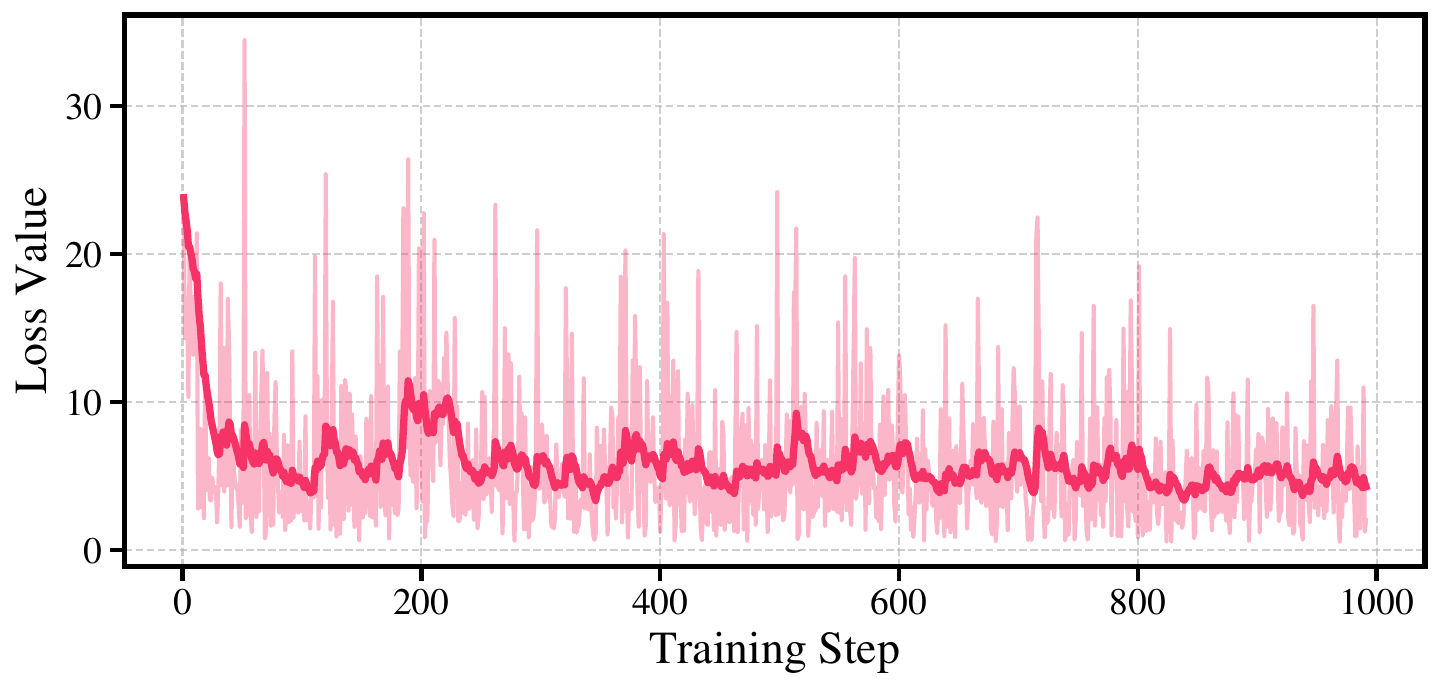}
    \end{minipage}
    \begin{minipage}{0.312\linewidth}
        \centering
        \includegraphics[width=\linewidth]{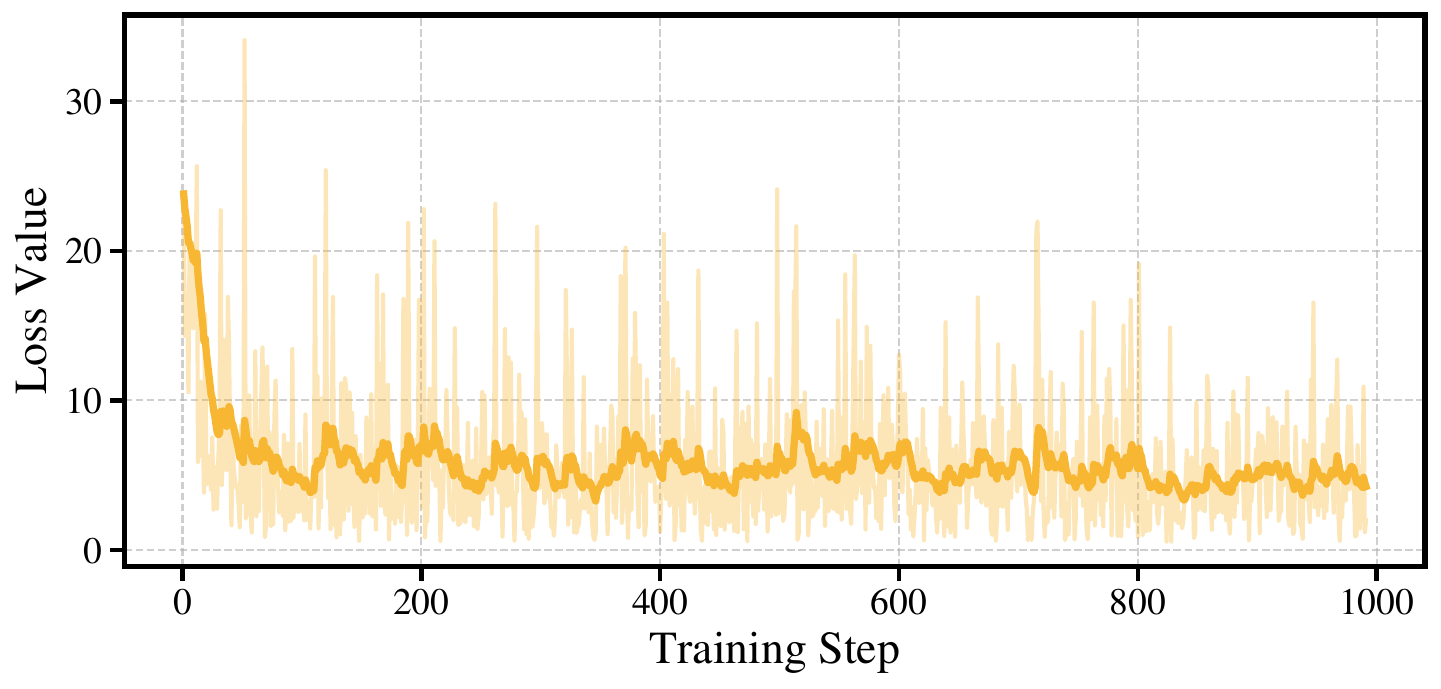}
    \end{minipage}
    \begin{minipage}{0.312\linewidth}
        \centering
        \includegraphics[width=\linewidth]{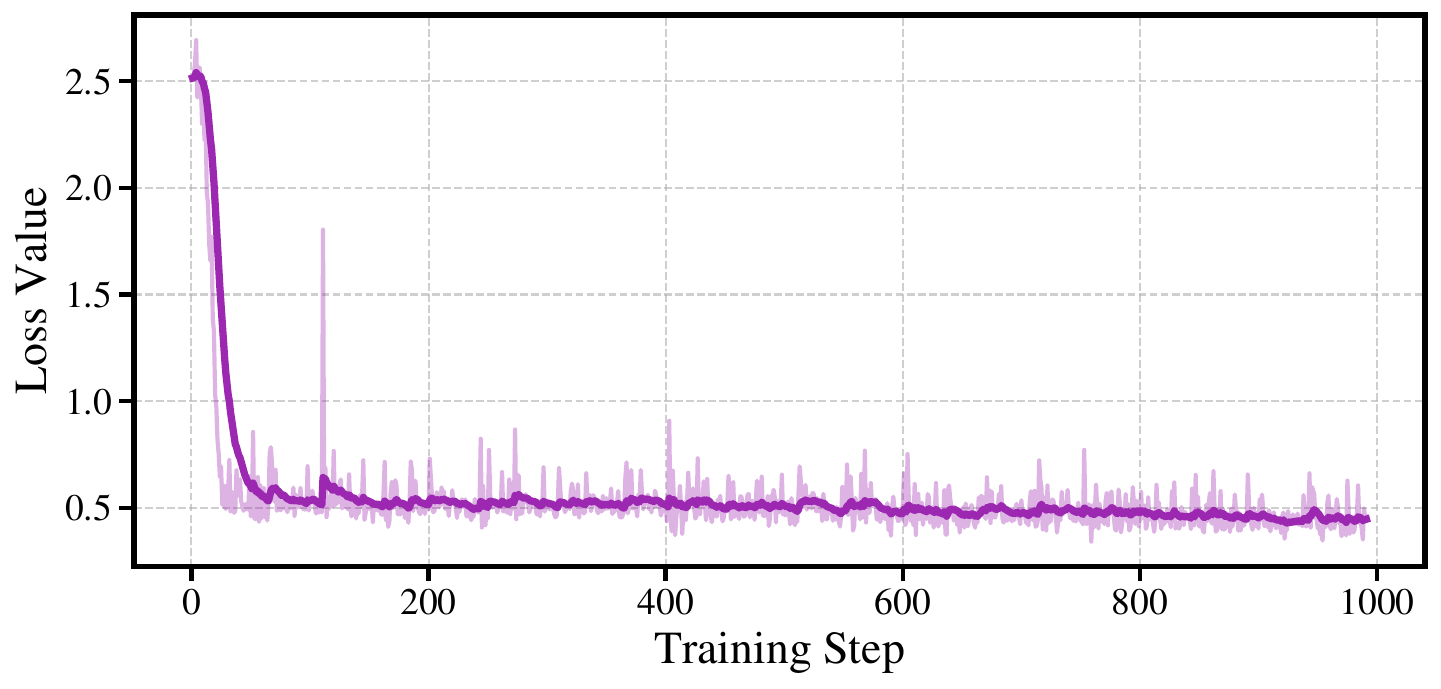}
    \end{minipage}\\[0.1em]

    \begin{minipage}{0.312\linewidth}
        \centering
        \includegraphics[width=\linewidth]{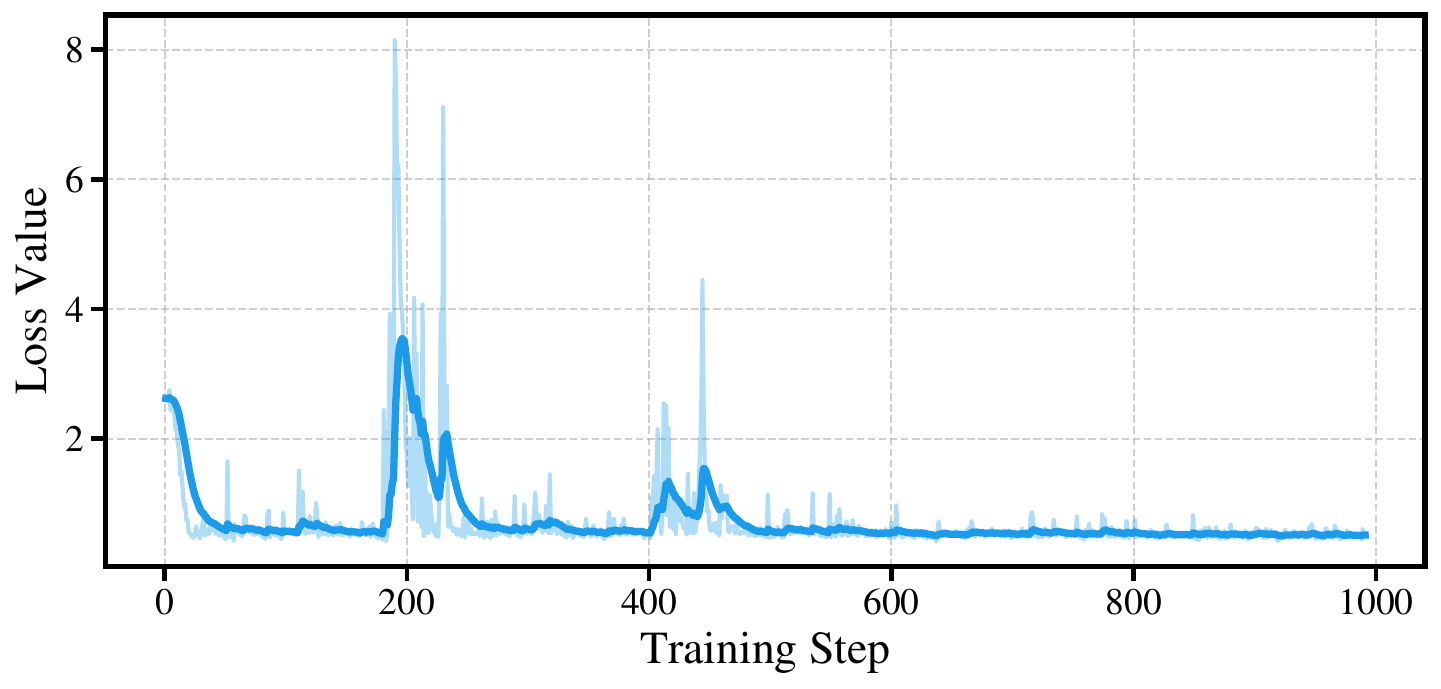}
    \end{minipage}
    \begin{minipage}{0.312\linewidth}
        \centering
        \includegraphics[width=\linewidth]{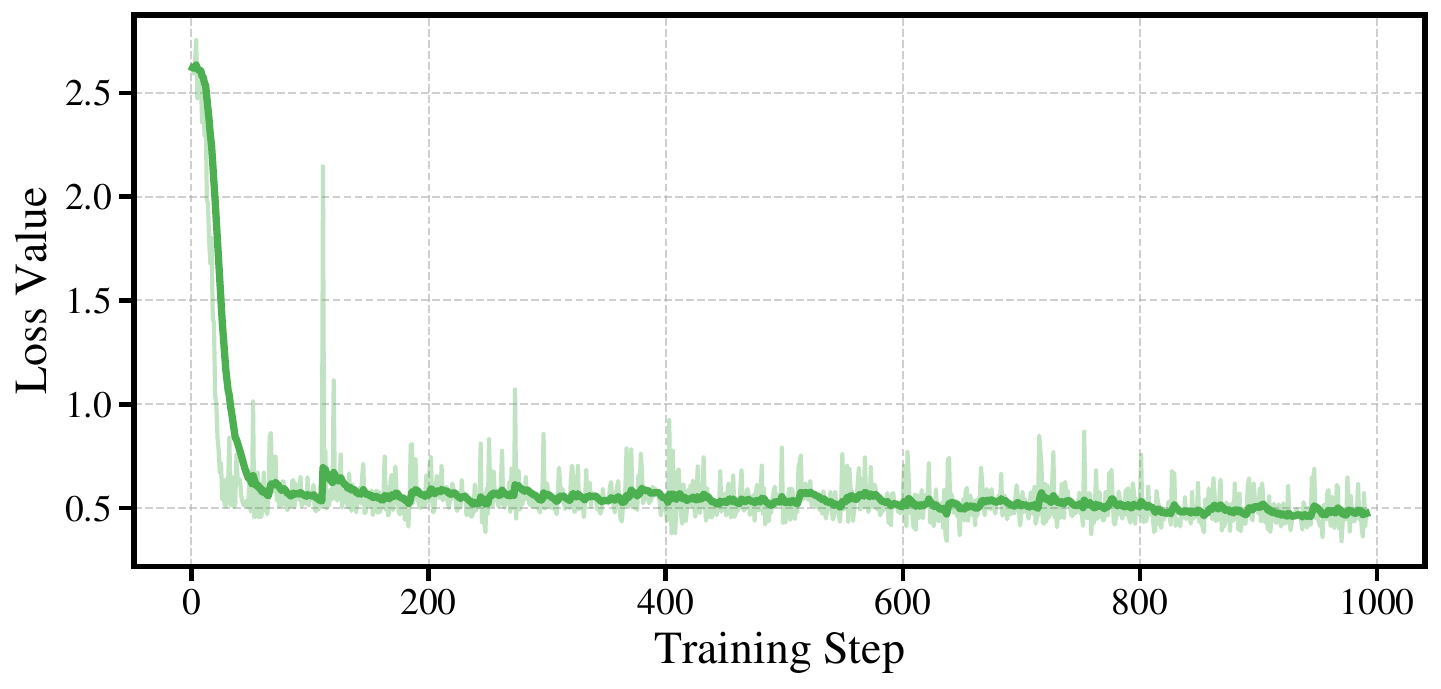}
    \end{minipage}
    \begin{minipage}{0.312\linewidth}
        \centering
        \includegraphics[width=\linewidth]{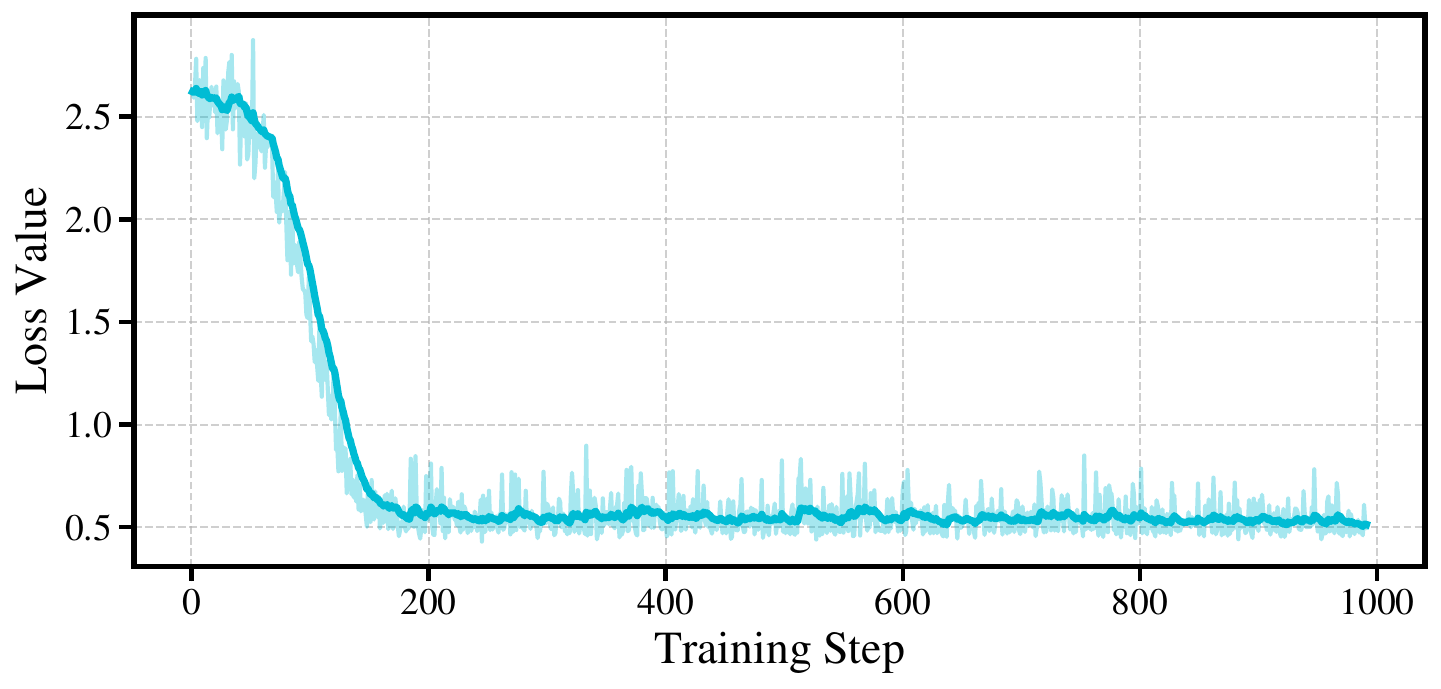}
    \end{minipage}

    \caption{Training loss trajectories for all experiments in the hyperparameter sweep.}
    \label{fig:train_loss_overlay}
\end{figure*}

\newcommand{\Triplet}[1]{%
  \subcaptionbox{Original}[.32\linewidth]{%
    \includegraphics[width=\linewidth]{fig/attention_sample_#1_orig.png}}%
  \hfill
  \subcaptionbox{Base Qwen2-VL-7B}[.32\linewidth]{%
    \includegraphics[width=\linewidth]{fig/attention_sample_#1_base_attn.png}}%
  \hfill
  \subcaptionbox{SFT Qwen2-VL-7B}[.32\linewidth]{%
    \includegraphics[width=\linewidth]{fig/attention_sample_#1_sft_custom.png}}%
}

\begin{table*}[htbp]
\centering
\scriptsize
\caption{Performance metrics of fine-tuned Qwen2-VL models with different training settings.}
\begin{tabular}{lccccccc}
\hline
\textbf{Model} & \textbf{Learning Rate (lr)} & \textbf{Coord $\boldsymbol{\lambda}$} & \textbf{BLEU-4} & \textbf{Language Modeling (LM) Loss} & \textbf{ROUGE-1} & \textbf{ROUGE-2} & \textbf{Mean Squared Error} \\
\hline
Qwen2-VL-7B\_ft & 1e-03 & 2e+01 & 0.65804 & 0.58734$\downarrow$ & 0.82091$\downarrow$ & 0.70300$\downarrow$ & 2337.53312$\downarrow$ \\
Qwen2-VL-7B\_ft & 1e-03 & 1e-01 & 0.65804 & 0.58317$\downarrow$ & 0.82000$\downarrow$ & 0.70200$\downarrow$ & 2374.61250$\downarrow$ \\
Qwen2-VL-7B\_ft & 1e-03 & 1e-04 & 0.65804 & 0.58103$\downarrow$ & 0.81909$\downarrow$ & 0.70100$\downarrow$ & 2611.39500$\downarrow$ \\
Qwen2-VL-7B\_ft & 5e-04 & 2e+01 & 0.65804 & 0.56108$\downarrow$ & 0.82091$\downarrow$ & 0.70300$\downarrow$ & 2303.35250$\uparrow$ \\
\textbf{Qwen2-VL-7B\_ft} & \textbf{5e-04} & \textbf{1e-01} & \textbf{0.65804} & \textbf{0.53045} & \textbf{0.82455} & \textbf{0.70700} & \textbf{2322.81000} \\
Qwen2-VL-7B\_ft & 5e-04 & 1e-04 & 0.65804 & 0.52957$\uparrow$ & 0.82091$\downarrow$ & 0.70300$\downarrow$ & 2382.19187$\downarrow$ \\
Qwen2-VL-7B\_ft & 1e-04 & 2e+01 & 0.65804 & 0.54917$\downarrow$ & 0.82182$\downarrow$ & 0.70400$\downarrow$ & 2446.92312$\downarrow$ \\
Qwen2-VL-7B\_ft & 1e-04 & 1e-01 & 0.65804 & 0.54627$\downarrow$ & 0.82273$\downarrow$ & 0.70500$\downarrow$ & 2360.06750$\downarrow$ \\
Qwen2-VL-7B\_ft & 1e-04 & 1e-04 & 0.65804 & 0.54107$\downarrow$ & 0.82273$\downarrow$ & 0.70500$\downarrow$ & 2370.25688$\downarrow$ \\
Qwen2-VL-7B\_ft & 5e-05 & 2e+01 & 0.65804 & 0.55185$\downarrow$ & 0.82091$\downarrow$ & 0.70300$\downarrow$ & 2316.67688$\uparrow$ \\
Qwen2-VL-7B\_ft & 5e-05 & 1e-01 & 0.65804 & 0.55523$\downarrow$ & 0.82182$\downarrow$ & 0.70400$\downarrow$ & 2387.05000$\downarrow$ \\
Qwen2-VL-7B\_ft & 5e-05 & 1e-04 & 0.65804 & 0.55235$\downarrow$ & 0.82182$\downarrow$ & 0.70400$\downarrow$ & 2412.89187$\downarrow$ \\
Qwen2-VL-7B\_ft & 1e-05 & 2e+01 & 0.65804 & 0.58621$\downarrow$ & 0.81818$\downarrow$ & 0.70000$\downarrow$ & 2303.76625$\uparrow$ \\
Qwen2-VL-7B\_ft & 1e-05 & 1e-01 & 0.65804 & 0.58742$\downarrow$ & 0.82000$\downarrow$ & 0.70100$\downarrow$ & 2508.41000$\downarrow$ \\
Qwen2-VL-7B\_ft & 1e-05 & 1e-04 & 0.65804 & 0.58727$\downarrow$ & 0.81909$\downarrow$ & 0.70100$\downarrow$ & 2371.03937$\downarrow$ \\
Qwen2-VL-2B (HF) & N/A & N/A & 0.01641$\downarrow$ & 2.28136$\downarrow$ & 0.17672$\downarrow$ & 0.05450$\downarrow$ & 566751.88000$\downarrow$ \\
Qwen2-VL-7B (HF) & N/A & N/A & 0.00274$\downarrow$ & 2.03562$\downarrow$ & 0.32676$\downarrow$ & 0.13376$\downarrow$ & 566751.88000$\downarrow$ \\
\hline
\end{tabular}
\vspace{2mm}
\parbox{0.95\textwidth}{\footnotesize
\textbf{Notes.} Arrows denote comparison against the selected best model (\textbf{Qwen2-VL-7B\_ft, lr=5e-04, coord=1e-01}): $\uparrow$ means better, $\downarrow$ means worse; no arrow means equal.
Higher is better for BLEU-4/ROUGE; lower is better for LM Loss/MSE.
}
\label{tab:qwen_eval_all}
\end{table*}

\subsubsection{Annotated Dataset}
We release an annotated subset of the BDD100K dataset\footnote{\url{https://huggingface.co/datasets/chdw98/llamafactory_bdd100k_dataset_1000_xy}}, consisting of 1{,}000 manually annotated images. Each example contains an RGB image and a short two-turn chat record stored under the field \texttt{messages}: a user turn that includes the image and a prompt (``\texttt{<image>} Which area should we pay more attention to in the current autonomous driving scenario?''), followed by an assistant turn that indicates the human-annotated hazard center using pixel coordinates, e.g., ``The area around \((x,y)\) should be paid more attention to.'' The original image resolution is \(1{,}280 \times 720\) pixels, and the \((x,y)\) labels correspond to this resolution and are included as part of the assistant response.\footnote{See the dataset viewer for schema details, including modalities (Image, Text), storage format (Parquet), split size (1k rows), and example rows illustrating the two-message structure and coordinate strings.}

For experiments, we use this 1{,}000-sample subset with a 90/10 split for training and validation. Images are resized while preserving aspect ratio such that the longer side is 512 pixels. The hazard coordinates are normalized to \([0,1]^2\) using the original image width and height and later denormalized to pixel space for evaluation.

We fine-tune Qwen2-VL-Instruct on this dataset using parameter-efficient LoRA and the coordinate-prediction head described in Sec.~\ref{sec:insight}, optimizing the multi-task loss \(\mathcal{L}_{\text{total}}\) in Eq.~\eqref{eq:multi_task_loss}.

\subsection{Evaluation Metrics}
The fine-tuned model's performance is evaluated using several core indicators, including BLEU-4, ROUGE-L, ROUGE-1, ROUGE-2, and Mean Squared Error (MSE). 
\subsubsection{BLEU-4 Metric}
The BLEU-4 measures the precision of up to 4-gram overlaps between a fine-tuned vision-language model's generated text and reference outputs, making it useful for tasks like image captioning comparison, ROUGE-1 captures the unigram-level overlap, indicating how well the model covers key information in text summarization tasks, ROUGE-2 extends this to bigrams, offering a finer-grained measure of contextual coverage for summarizing or describing visual content, and ROUGE-L uses the longest common subsequence to evaluate the sequence-level match, emphasizing overall recall of critical information in the model’s output. The MSE calculation function is shown below:
\begin{equation}
    \text{MSE} = \frac{1}{n} \sum_{i=1}^n \left( (x_i - x_i^{\text{true}})^2 + (y_i - y_i^{\text{true}})^2 \right)
\end{equation}
where $x_i$ and $y_i$ represent the prediction from the fine-tuned Qwen2-VL and the $x_i^{\text{true}}$ and $y_i^{\text{true}}$ represent the human-annotation ground truth pixels location.


\subsection{Experiment and Training Details}
\subsubsection{LoRA configuration}
We apply LoRA to the attention projection modules \{q\_proj, k\_proj, v\_proj, o\_proj\} with rank $r=8$, scaling factor $\alpha=16$, dropout rate $0.05$, and no bias, using AdamW as the optimizer.

\subsubsection{Coordinate Head and Objective}
On top of the pooled multimodal backbone states, we add a small MLP head
$d \rightarrow d/2 \rightarrow (H{\times}W)$ with $H{=}16$ and $W{=}16$. As in Sec.~\ref{sec:insight}, the head produces a heatmap $A \in \mathbb{R}^{H \times W}$ via softmax, and the predicted point $(\hat{x}, \hat{y})$ is given by the soft-argmax expectation over a normalized grid, yielding $(\hat{x},\hat{y}) \in [0,1]^2$. 

We use the multi-task loss
$\mathcal{L}_{\text{total}} \;=\; \lambda_{\text{text}}\,\mathcal{L}_{\text{text}} \;+\; \lambda_{\text{coord}}\,\mathcal{L}_{\text{coord}}$,
with $\lambda_{\text{text}}=1.0$ and $\lambda_{\text{coord}}=10^{-1}$. Besides these training losses, we also log a denormalized pixel-level MAE by mapping $(\hat{x},\hat{y})$ back to the original image resolution.

\subsubsection{Training Schedule and Logging}

All experiments use a per-device batch size of 1 with no gradient accumulation and are trained for 3 epochs under a cosine learning-rate schedule with a warmup ratio of 0.1. The peak learning rate is swept over $1\times10^{-5}$, $5\times10^{-5}$, $1\times10^{-4}$, $5\times10^{-4}$, and $1\times10^{-3}$. The coordinate loss weight $\lambda$ is varied over $1\times10^{-4}$, $1\times10^{-3}$, $1\times10^{-2}$, $1\times10^{-1}$, 1, 5, 10, and 20.

Validation is performed at the end of every epoch. Training metrics are logged at each step to Weights \& Biases, and gradient and parameter norms are recorded every 100 steps through a custom callback. No intermediate checkpoints are saved during training; only the final LoRA adapter and processor are stored.

The random seed is fixed at 42. Training uses bf16 precision with 4-bit NF4 quantization and the paged AdamW 8-bit optimizer, together with gradient checkpointing. Each configuration is assigned a unique run name, and all metrics are summarized in a CSV file. Due to space constraints, Table~\ref{tab:qwen_eval_all} reports a representative subset of configurations.

\begin{figure*}[ht]
  \centering

  \begin{subfigure}[t]{0.49\textwidth}\centering\caption{Urban Scenario 1}\Triplet{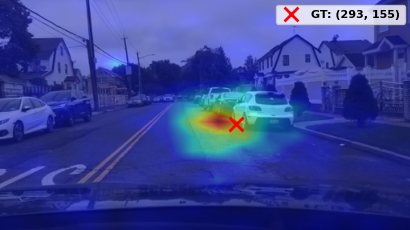}\end{subfigure}\hfill
  \begin{subfigure}[t]{0.49\textwidth}\centering\caption{Urban Scenario 2}\Triplet{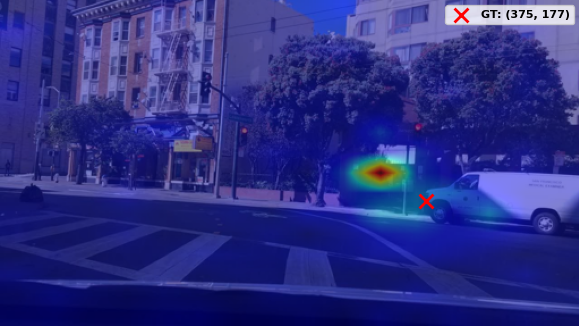}\end{subfigure}\hfill

  \vspace{0.8em}
  \begin{subfigure}[t]{0.49\textwidth}\centering\caption{Urban Scenario 3}\Triplet{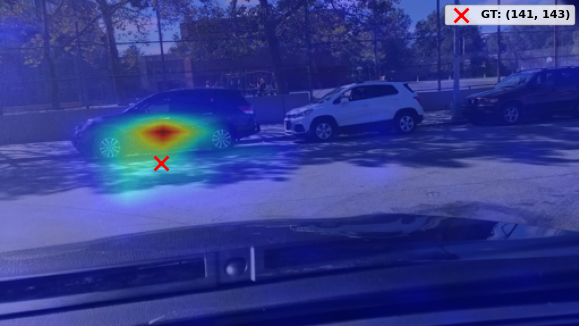}\end{subfigure}
  \begin{subfigure}[t]{0.49\textwidth}\centering\caption{Highway Scenario 1}\Triplet{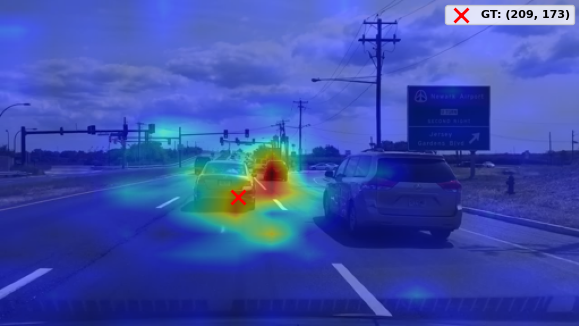}\end{subfigure}
  \vspace{0.8em}

  \begin{subfigure}[t]{0.49\textwidth}\centering\caption{Highway Scenario 2}\Triplet{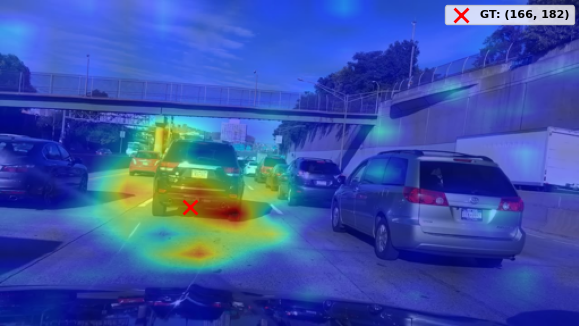}\end{subfigure}\hfill
  \begin{subfigure}[t]{0.49\textwidth}\centering\caption{Highway Scenario 3}\Triplet{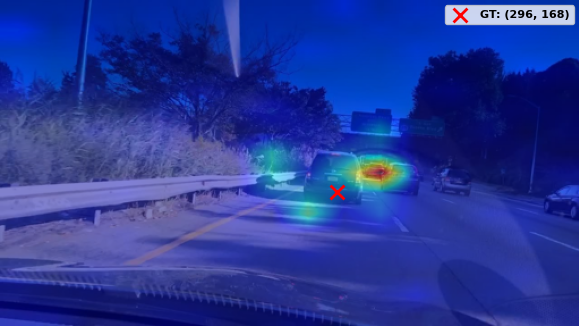}\end{subfigure}\hfill

  \caption{Original image and visualization of attention maps for the base and SFT models.}
  \label{fig:attn_3x3}
\end{figure*}

\subsubsection{Training Details}

To examine hyperparameter sensitivity, we analyze the interaction between the learning rate and the coordinate loss weight over the grid defined above. All configurations share identical training settings and differ only in these two factors.

Figure~\ref{fig:train_loss_overlay} presents the training loss curves of six representative configurations arranged in a $2\times3$ layout. From left to right and top to bottom, the settings correspond to learning rate $10^{-3}$ with $\lambda=20$, learning rate $5\times10^{-4}$ with $\lambda=20$, learning rate $5\times10^{-4}$ with $\lambda=10^{-4}$, learning rate $10^{-3}$ with $\lambda=10^{-1}$, learning rate $5\times10^{-4}$ with $\lambda=10^{-1}$, and learning rate $10^{-5}$ with $\lambda=10^{-1}$.

Large coordinate weights such as $\lambda=20$ introduce significant oscillations and instability, particularly when combined with a high learning rate of $10^{-3}$. Reducing the learning rate to $5\times10^{-4}$ improves stability but still results in elevated steady-state loss.

With moderate scaling $\lambda=10^{-1}$, optimization becomes substantially more stable. The configuration using learning rate $5\times10^{-4}$ and $\lambda=10^{-1}$ achieves rapid early decay and smooth convergence. Increasing the learning rate to $10^{-3}$ leads to occasional spikes, whereas decreasing it to $10^{-5}$ slows early convergence despite stable behavior.

When the coordinate weight is very small at $10^{-4}$, the loss curve remains smooth; however, evaluation metrics indicate slightly weaker localization performance due to insufficient emphasis on coordinate supervision.

Overall, excessively large coordinate weights destabilize optimization, while very small learning rates slow convergence without measurable gains. The setting with learning rate $5\times10^{-4}$ and $\lambda=10^{-1}$ provides the best balance between stability and performance and is adopted in subsequent experiments.

\subsection{Results and Analysis}

\subsubsection{Quantitative Analysis}

Table~\ref{tab:qwen_eval_all} reports the performance under different learning rates and coordinate-loss weights $\lambda_{\text{coord}}$. Across all fine-tuned Qwen2-VL-7B variants, BLEU-4 remains constant at 0.65804, while ROUGE-1/2 stay within a narrow range ($\simeq$0.82/0.70), indicating stable language generation quality. In contrast, LM loss and MSE show clearer sensitivity to hyperparameter choices.

The configuration $(\textit{lr}=5\times10^{-4},\, \lambda_{\text{coord}}=10^{-1})$ achieves the best overall trade-off and is selected as the reference model. It yields the lowest LM loss (0.53045), the highest ROUGE-1/2 (0.82455/0.70700), and competitive coordinate accuracy.

With $\lambda_{\text{coord}}=2\times10^{1}$, optimization becomes less balanced: although MSE slightly decreases, LM loss increases and ROUGE scores decline, suggesting overemphasis on coordinate regression. Conversely, $\lambda_{\text{coord}}=10^{-4}$ marginally lowers LM loss (0.52957) but degrades ROUGE and increases MSE, indicating weakened spatial supervision.

Across learning rates, $5\times10^{-4}$ consistently performs best. A larger rate ($10^{-3}$) increases LM loss to $\simeq$0.58–0.59, while a very small rate ($10^{-5}$) leads to higher LM loss ($\simeq$0.587) and worse MSE, implying underfitting. Compared with off-the-shelf Qwen2-VL baselines, fine-tuning substantially improves BLEU and ROUGE, reduces LM loss from above 2.0 to below 0.6, and decreases MSE from $5.6\times10^{5}$ to the $10^{3}$ range, confirming the necessity of task-specific adaptation.






\subsubsection{Qualitative Analysis}
To further investigate how supervised fine-tuning (SFT) influences the model's visual grounding capability, we present a qualitative comparison of the attention maps generated by the base and SFT Qwen2-VL-7B models, as shown in Fig.~\ref{fig:attn_3x3}. Each row corresponds to one driving scenario, displaying the original image, the base model attention, and the SFT-enhanced attention heatmap.

In urban driving scenes (Samples~1 and~2), the base model exhibits dispersed and sometimes misplaced attention, occasionally focusing on irrelevant areas such as the sky or background objects. In contrast, the SFT model demonstrates a sharper and more concentrated focus near the ground-truth (GT) location, effectively aligning with salient traffic participants such as the leading vehicle or nearby road users. This indicates that SFT enhances spatial grounding by aligning multimodal reasoning with visual saliency.

In highway scenes (Sample~3), where vehicles are densely packed and motion cues are less explicit, the base model tends to distribute attention broadly across multiple vehicles. The SFT model, however, correctly emphasizes the primary vehicle in the lane, showing stronger context-awareness and reduced background distraction. This improvement highlights the benefit of SFT in refining the model’s attention to task-relevant visual regions, especially under complex spatial layouts.






%% file: sec/conclusion.tex
\section{Conclusion}

This study demonstrates the effectiveness of fine-tuned vision-language models (VLMs), such as Qwen2-VL, in enhancing safety-critical scene understanding and contextual hazard reasoning for autonomous driving. By bridging visual and textual modalities, INSIGHT improves risk-aware perception and enables more accurate hazard region localization and descriptive reasoning. Experimental results show consistent gains in coordinate prediction and language generation compared to baseline models, reflecting stronger alignment between multimodal representations and safety-oriented scene interpretation. INSIGHT provides a scalable framework for integrating multimodal learning into autonomous systems to strengthen safety awareness and interpretability in complex driving environments. Future work will investigate real-time deployment, broader dataset generalization, integration with CARLA simulators, and validation in diverse real-world scenarios.